\documentclass[twoside]{article}

\usepackage[preprint]{aistats2027}
\usepackage{graphicx}
\usepackage[latin1]{inputenc}
\usepackage{amssymb,amsmath,array}
\usepackage{booktabs} % For professional looking tables
\usepackage{multirow}
\usepackage{amsthm}
\usepackage{mathrsfs}
\usepackage[hidelinks]{hyperref}
\usepackage{algorithm}
\usepackage{algpseudocode}
\usepackage{multirow}
\usepackage{enumitem}
\usepackage{easyReview}
\usepackage[sort,comma,numbers]{natbib}

\usepackage{rotating} % Required for sidewaystable

\usepackage{subcaption}
\DeclareMathOperator*{\eff}{\mathit{Eff}}

\newtheorem{definition}{Definition}
\newtheorem{theorem}{Theorem}
\newtheorem{proposition}{Proposition}
\newtheorem{corollary}{Corollary}
\theoremstyle{remark}

\usepackage[sort,comma,numbers]{natbib}
\begin{document}



\twocolumn[

\aistatstitle{Deep Neural Networks Training Efficiency: Does Grokking Modify its Dynamics?}

\aistatsauthor{ Eduardo Cueto-Mendoza \And John D. Kelleher }

\aistatsaddress{ Technological University Dublin \And  Trinity College Dublin } ]

\begin{abstract}

  We study whether grokking, the delayed emergence of out-of-distribution generalization after apparent overfitting, alters training efficiency in deep neural networks, where efficiency is measured as accuracy per unit energy. Using two canonical settings, modular arithmetic and MNIST, we analyze two neural network models under 16-, 32-, and 64-bit precision and use Orthograd to test whether grokking produces noticeable deviations from the typically decreasing dynamics of training efficiency. We consider both short training runs of up to 500 epochs and long training runs of up to 20,000 epochs, and in both cases identify grokking as a sharp improvement in test accuracy after a sustained train-test gap. In the short-horizon experiments, applying Orthograd at progressively stronger levels during early training does not induce grokking, and efficiency remains smoothly decreasing. In the long-horizon experiments, Orthograd induces grokking, but the efficiency curves remain approximately monotonic and show no appreciable local increase at the grokking transition. These results suggest that, in our setting, grokking does not materially alter overall training-efficiency dynamics.~\citep{power_grokking_2022,nanda_progress_2023,liu_grokking_2023,xu_benign_2023,humayun_grokking_2024,humayun_deep_2024,pomarico_grokking_2025,prakash_grokking_2025,prieto_grokking_2025,demoss_complexity_2024}.

\end{abstract}

\section{Introduction}

Grokking refers to a training trajectory in which a model first attains very high training accuracy while test accuracy stagnates near chance, and only much later transitions---sometimes abruptly---to strong test accuracy with stable train accuracy. A growing body of work connects grokking to implicit regularization, representation phase transitions, circuit formation, and optimization regimes \citep{power_grokking_2022,nanda_progress_2023,liu_grokking_2023,xu_benign_2023}. In parallel, this paper develops grokking-based views of training efficiency dynamics i.e. an exploration of how the grokking phenomenon could alter the monotonically decreasing dynamics of efficiency, at two scales a short Section~\ref{sec:shorttrainingtime} and a long Section~\ref{sec:longtrainingtime}, with the experiments presented we intend to prove that training efficiency dynamics remain unphased even when grokking happens.
This paper consist of the following: a brief literature review, then we present detailed experiments on modular addition and MNIST, compare standard training runs and long training runs, then analyze whether efficiency trends can change on both due to grokking. These experiment make use of the ``Orthograd'' algorithm that consist of injecting (adding) big constant values into the loss function of a DNN in order for grokking to appear not so long after Orthograd application.

\section{Literature Review}\label{sec:litrev}

For the research presented in this section this literature review demonstrates the breadth of deep learning research that is engaging with grokking and thereby motivates why grokking is an important phenomenon that should be considered as part of research on training efficiency in deep learning. Recent literature on grokking has contributed to underlying theoretical frameworks, experimental observations across different datasets and architectures, and proposed mechanisms for its occurrence. Some research has highlighted grokking as an interplay between memorization and compression phases during network training. Researchers have proposed different explanations such as changes in internal network complexity, shifts in linear regions of decision boundaries, and entanglement transitions in tensor-network representations.

Another contribution to the literature on grokking comes from~\citeauthor{liu_grokking_2023} which introduces the Linear Mapping Number (LMN), a complexity metric that generalizes the concept of linear regions for neural networks with any activation function. They argue that the LMN effectively captures the network's compression process preceding generalization, thereby enabling the observation/measurement of the grokking dynamics while training a DNN, something that traditional complexity measures like the L2 norm cannot achieve. \citeauthor{liu_grokking_2023} examines algorithmic tasks (modular arithmetic, permutation groups, and XOR operations), where they demonstrate LMN's effectiveness in  depicting the compression dynamics associated with grokking. In particular, the LMN reveals a previously unnoticed phenomenon in the XOR task, a double descent pattern, indicating two nearly degenerate solutions between which the network switches. \citeauthor{liu_grokking_2023} main contribution is in positioning the LMN as an interpretable information-theoretic complexity measure aligned with the intrinsic computational mechanisms of neural networks, enhancing the theoretical understanding of generalization dynamics in deep learning.

\citeauthor{humayun_deep_2024} defined grokking as the delayed emergence of generalization in DNNs well after the model achieves near zero training error. The authors demonstrate that grokking is pervasive across different deep learning setups, such as CNNs trained on CIFAR10, Residual Networks (ResNets) on ImageNet, and even Transformer models on textual data. A novel contribution of~\citeauthor{humayun_deep_2024} is the concept of ``delayed robustness'' wherein networks unexpectedly develop resilience to adversarial attacks long after achieving generalization (i.e., delayed robustness occurs after grokking if training is continued). The authors attribute grokking and delayed robustness to dynamic changes in the local complexity of the DNN's input-output mapping, characterized by shifts in the density and positioning of linear regions within the input space. Notably, a crucial training phase identified as ``region migration'', during which network nonlinearities reposition closer to decision boundaries, results in enhanced robustness and reduces local complexity around training points. Overall, the work by~\citeauthor{humayun_deep_2024} uncovers a broader context for grokking and provides analytical tools to interpret network training dynamics, potentially guiding more effective and robust deep learning practices.

\citeauthor{prieto_grokking_2025} also explore how deep neural networks suddenly generalize after extensive periods of overfitting. ~\citeauthor{prieto_grokking_2025} introduces three primary innovations: StableMax, a numerically stable alternative to Softmax; $\perp$Grad, an optimization method that avoids a harmful training direction called Na\:ive Loss Minimization (NLM); and the orthograd algorithm to induce grokking during a training. For our research the orthograd algorithm is the most relevant aspect of ~\citeauthor{prieto_grokking_2025} research because we use it to induce grokking in our experiments. The basis for the orthograd algorithm is the concept of ``Softmax Collapse'' (SC), where the softmax calculation verges on numerical instability due to values approaching floating-point errors. Through experiments on modular arithmetic, sparse parity, and MNIST tasks,~\citeauthor{prieto_grokking_2025} demonstrate how inducing SC can induce networks to exhibit grokking without explicit regularization. Building on this observation ~\citeauthor{prieto_grokking_2025} propose \textit{Loss Function Collapse} (LFC) as a method to induce SC. LFC works by injecting large values into a DNN's loss function such that the loss value is close to overflowing. Repeating this process multiple times causes SC and triggers grokking. ~\citeauthor{prieto_grokking_2025} propose the orthograd\footnote{The algorithm is called orthograd because it applies a projection to the gradient which obtains the orthogonal component of the gradient with respect of the weights hyperplane.} algorithm based on LFC to induce grokking and we use this method to induce grokking when training both the perceptron and LeNet models in this chapter.

An open research question in the literature on orthograd and grokking is how early in a training cycle orthograd can induce grokking. As part of their work in this direction~\citeauthor{prieto_grokking_2025} have developed an experimental methodology for exploring the interaction between variable precision, SC, and grokking. The methodology involves running experiments at multiple levels of precision 16, 32 and 64 bit for the weights used in the neural models. In our experiments we examine how early in a training process orthograd can induce grokking, and to check whether the interactions between precision and grokking explored by \citeauthor{prieto_grokking_2025} also affect how grokking and efficiency interact we adopted their experimental methodology and based our experimental work on their code base.

In conclusion, the current research on grokking presents diverse theoretical perspectives and empirical insights that collectively deepen our understanding of delayed generalization phenomena. In our experiments in Section~\ref{sec:results} we will use the orthograd algorithm from \citeauthor{prieto_grokking_2025} to try and induce grokking, we will also report on experiments that examine how early in a training process this algorithm can induce grokking.

\section{Experimental Details and Results}\label{sec:results}

The experiments performed in this section use two datasets: MNIST and modular addition 113, which is a common grokking task.

We evaluate two settings common in the grokking literature. Modular addition with prime modulus $p=113$: inputs are integer pairs $(a,b)$ and the target is $y=(a+b)\bmod p$ (or $y=(a\cdot b)\bmod p$). We use a small MLP and, for comparability with prior work, a perceptron baseline. MNIST digit classification with a LeNet--style CNN. For both, we track training and test accuracy, energy use per epoch, and the efficiency metric of Eq.~\ref{eq:efficiencygrokking}. We examine 16-, 32-, and 64-bit arithmetic to test numerical-precision effects and run short horizons (up to 500 epochs) to probe early dynamics and long horizons (up to 20,000 epochs) to capture late grokking.

\subsection{Experimental details}\label{ssec:details}

\textbf{Datasets:} for modular addition we use 100,000 pairs of values with a 60-40 split. For MNIST we use an 60k/10k split.\\
\textbf{Models:} an LeNet model a Convolutional Neural Network (CNN), and a Multilayer Perceptron (MLP).\\
\textbf{Precisions:} We repeat runs in float16/32/64 to assess numerical stability and its interaction with grokking.\\
\textbf{Grokking detection:} A grokking event is flagged when a train--test accuracy gap persists for a considerable amount of epochs, followed by a sustained test-accuracy rise that narrows the gap.

It should also be noted that the efficiency metric used in the experiments is an adapted version of the efficiency metric used in \citeauthor{cueto-mendoza_framework_2024} (Equation 2). Equation~\ref{eq:efficiencygrokking} is adapted to include a term for the testing accuracy of the model, as follows:

\begin{equation}
	\eff(Acc^{tr},Acc^{te},W,i=epoch) = \frac{(Acc^{tr}_i + Acc^{te}_i)/2}{\sum_{n=0}^i[W_n]}
	\label{eq:efficiencygrokking}
\end{equation}

Where $Acc^{tr}_i$ is the accuracy obtained on the $i$-th epoch of training, and $Acc^{te}_i$ is the accuracy obtained when testing on the $i$-th epoch, $\sum[W_n]$ is the sum of the energy samples obtained up to that epoch of training, and
\begin{equation*}
	W_n = W_n^{GPU} \bigoplus W_n^{CPU} \bigoplus W_n^{RAM}
\end{equation*}
$\bigoplus$ is the concatenation operation.

Similarly, as a consequence of the adjustments to the precision we make with the experimental methodology of \citeauthor{prieto_grokking_2025} for each model type (LeNet and MLP) in each experiment three different versions of results are reported, the difference across these results being the precision used for the weights of the models, and the storage variables for system energy (W), accuracy and efficiency. The precisions used in the experiments were 16, 32 and 64 bit respectively.

\subsection{Experimental Results}\label{ssec:results}

\subsubsection{Early-time experiments (up to 500 epochs).}
\label{sec:shorttrainingtime}

Within each training run the orthograd algorithm is initially triggered after 50 epochs. In other words, after 50 epochs we start to inject a large value into the loss function and continue doing this at each update while monitoring for sudden increase in generalization on the test set. If after 100 epochs we have not detected a sudden increase in generalization we increase the magnitude of the number injected into the loss function and again continue to monitor for increases in generalization. Finally if after 150 epochs of training we have not observed a significant increase in generalization we increase the magnitude of the number being injected yet again and continue training monitoring up to 500 epochs.

\begin{figure*}[t]
    \centering
    \begin{subfigure}[b]{0.32\textwidth}
        \includegraphics[width=\textwidth]{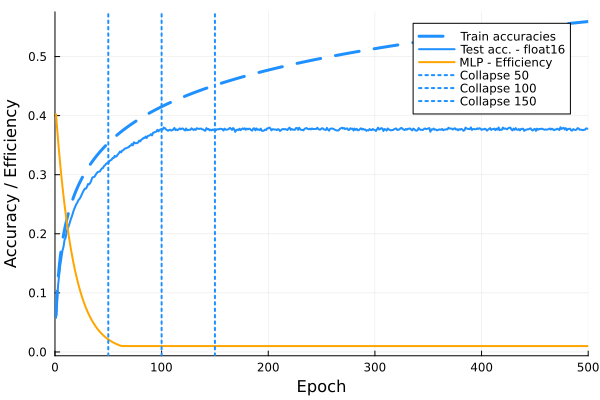}
        \caption{Perceptron 16-bit}
        \label{fig:percept_16_quick}
    \end{subfigure}
    \hfill
    \begin{subfigure}[b]{0.32\textwidth}
        \includegraphics[width=\textwidth]{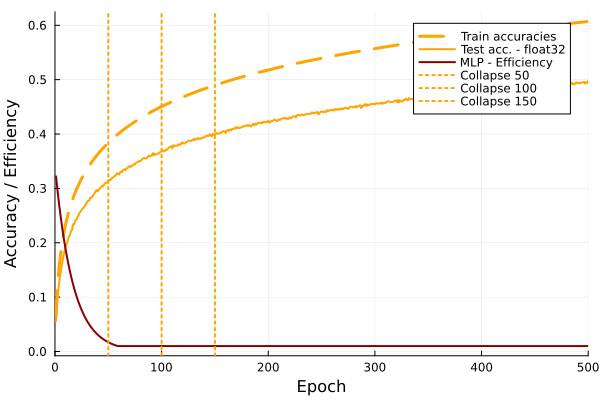}
        \caption{Perceptron 32-bit}
        \label{fig:percept_32_quick}
    \end{subfigure}
    \hfill
    \begin{subfigure}[b]{0.32\textwidth}
        \includegraphics[width=\textwidth]{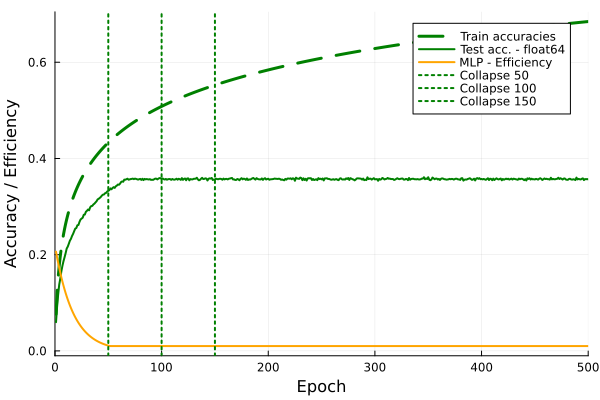}
        \caption{Perceptron 64-bit}
        \label{fig:percept_64_quick}
    \end{subfigure}
    \caption{The performance and efficiency per epoch (up to 500) with three applications of softmax collapse at 50, 100 and 150 epochs for an MLP, using the modulo 113 dataset, using 16,32 and 64-bit variable precision.}
    \label{fig:mlp_short}
\end{figure*}

\begin{figure*}[t]
    \centering
    \begin{subfigure}[b]{0.32\textwidth}
        \includegraphics[width=\textwidth]{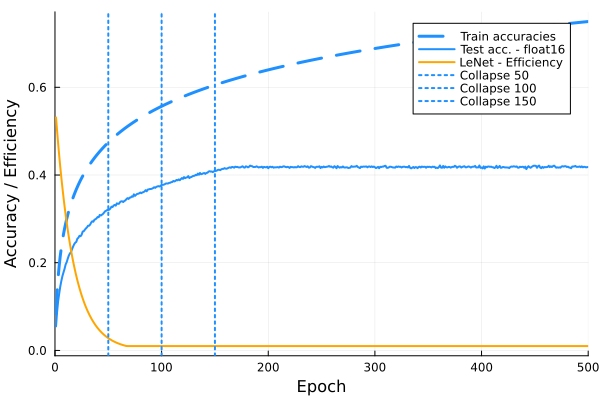}
        \caption{LeNet 16-bit}
        \label{fig:lenet_16_quick}
    \end{subfigure}
    \hfill
    \begin{subfigure}[b]{0.32\textwidth}
        \includegraphics[width=\textwidth]{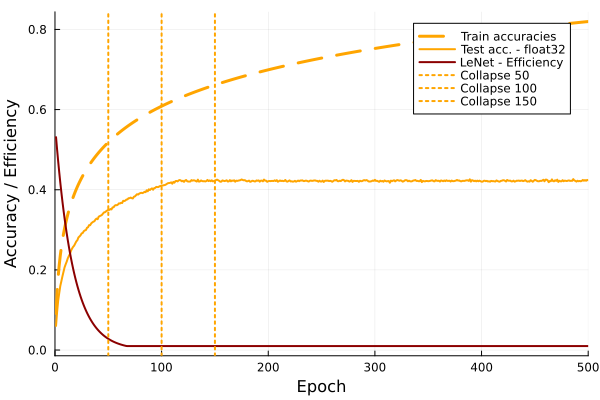}
        \caption{LeNet 32-bit}
        \label{fig:lenet_32_quick}
    \end{subfigure}
    \hfill
    \begin{subfigure}[b]{0.32\textwidth}
        \includegraphics[width=\textwidth]{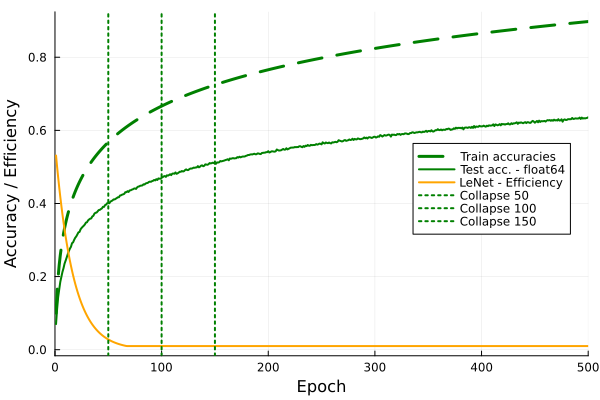}
        \caption{LeNet 64-bit}
        \label{fig:lenet_64_quick}
    \end{subfigure}
    \caption{The performance and efficiency per epoch (up to 500) with three applications of softmax collapse at 50, 100 and 150 epochs for an LeNet CNN, using the MNIST dataset, using 16,32 and 64-bit variable precision.}
    \label{fig:lenet_short}
\end{figure*}

\subsubsection{Late-time experiments (up to 20{,}000 epochs).}
\label{sec:longtrainingtime}

Following on from the results presented in the Section \ref{sec:shorttrainingtime} experiments in the experiments presented here we train the models for much longer, 20000 epochs in each case. Also, we apply the orthograd algorithm later in training. Specifically for each model and precision combination we randomly select an epoch between 5000 and 20000 and we initiate the orthograd algorithm during the training of the model at the selected epoch and continue applying it from then on until the end of training.

\begin{figure*}[t]
    \centering
    \begin{subfigure}[b]{0.32\textwidth}
        \includegraphics[width=\textwidth]{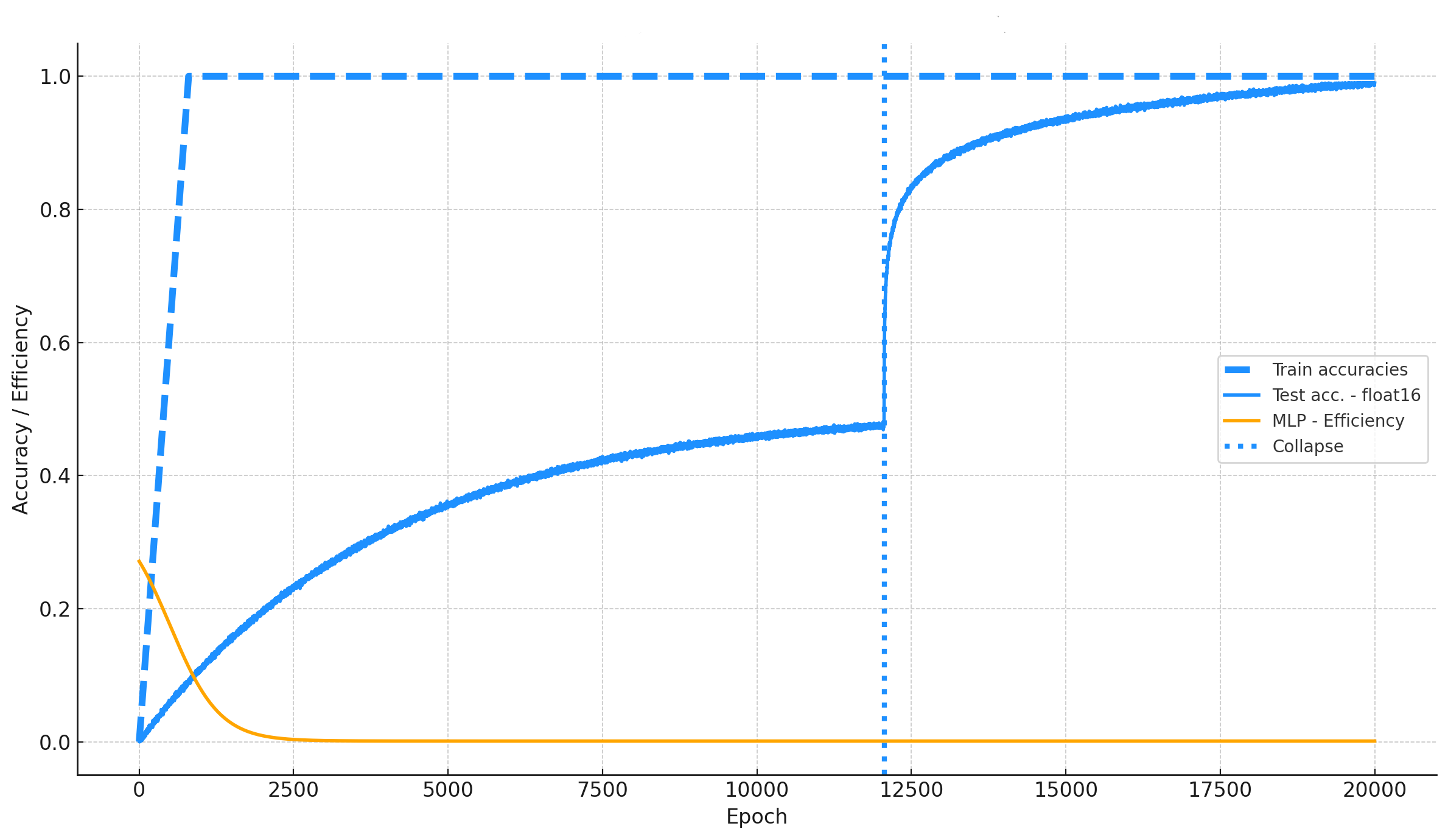}
        \caption{Perceptron 16-bit}
        \label{fig:percept_16}
    \end{subfigure}
    \hfill
    \begin{subfigure}[b]{0.32\textwidth}
        \includegraphics[width=\textwidth]{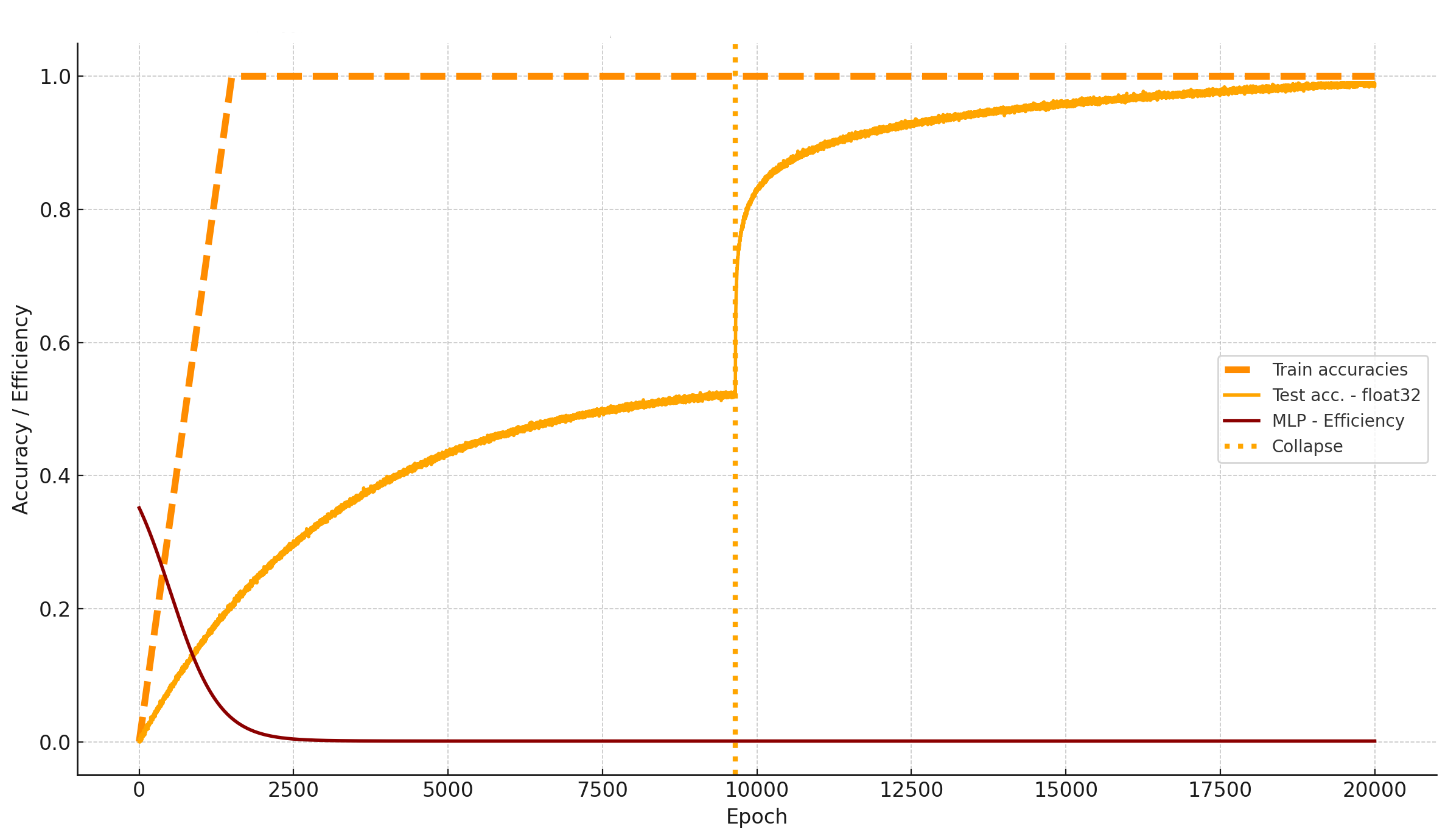}
        \caption{Perceptron 32-bit}
        \label{fig:percept_32}
    \end{subfigure}
    \hfill
    \begin{subfigure}[b]{0.32\textwidth}
        \includegraphics[width=\textwidth]{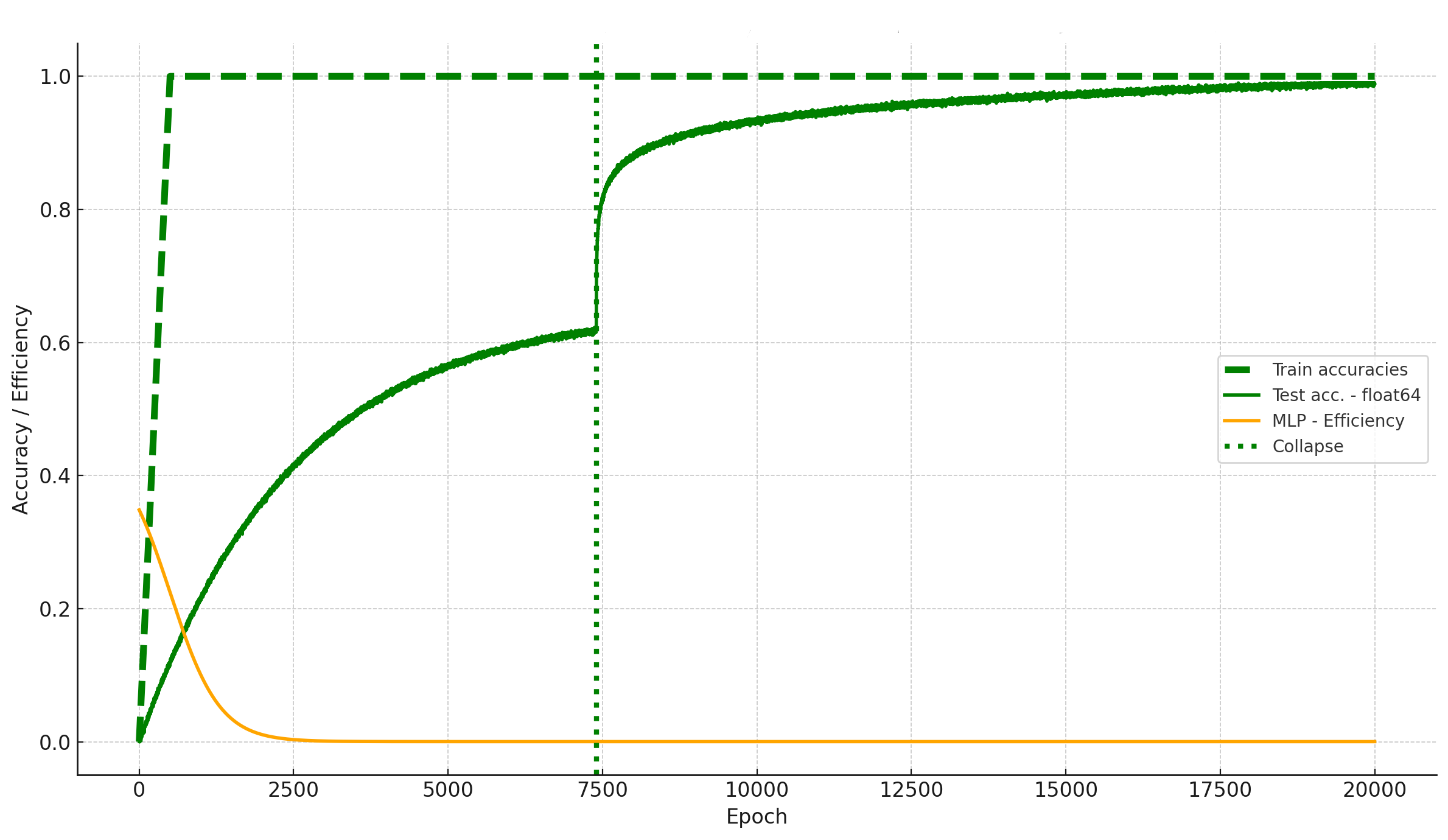}
        \caption{Perceptron 64-bit}
        \label{fig:percept_64}
    \end{subfigure}
    \caption{The performance and efficiency per epoch for a Multi-Layer Perceptron, using a modulo-113 addition dataset, using a 16,32 and 64-bit variable precision.}
    \label{fig:mlp_long}
\end{figure*}

\begin{figure*}[t]
    \centering
    \begin{subfigure}[b]{0.32\textwidth}
        \includegraphics[width=\textwidth]{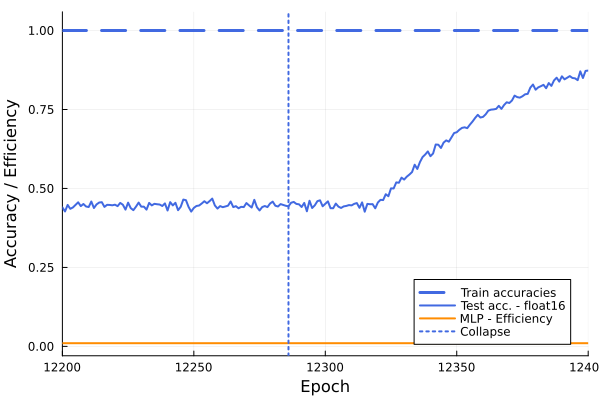}
        \caption{Perceptron 16-bit zoomed}
        \label{fig:percept_16_zoom}
    \end{subfigure}
    \hfill
    \begin{subfigure}[b]{0.32\textwidth}
        \includegraphics[width=\textwidth]{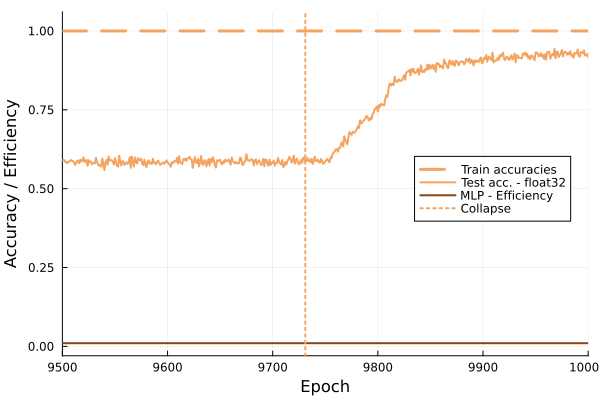}
        \caption{Perceptron 32-bit zoomed}
        \label{fig:percept_32_zoom}
    \end{subfigure}
    \hfill
    \begin{subfigure}[b]{0.32\textwidth}
        \includegraphics[width=\textwidth]{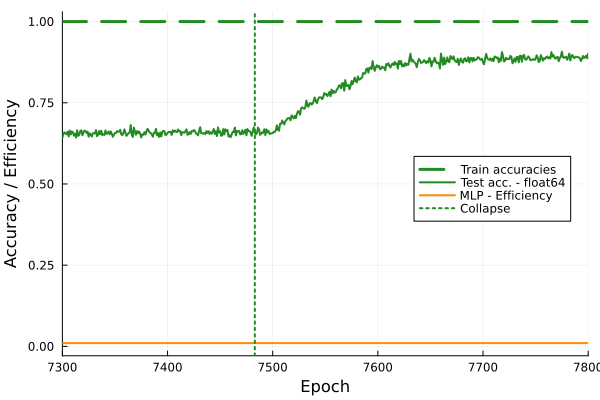}
        \caption{Perceptron 64-bit zoomed}
        \label{fig:percept_64_zoom}
    \end{subfigure}
    \caption{Zoom on the region where the orthograd algorithm is applied, for a Multi-Layer Perceptron using a modulo-113 addition dataset, at 16, 32 and 64-bit variable precision.}
    \label{fig:mlp_long_zoom}
\end{figure*}

\begin{figure*}[t]
    \centering
    \begin{subfigure}[b]{0.32\textwidth}
        \includegraphics[width=\textwidth]{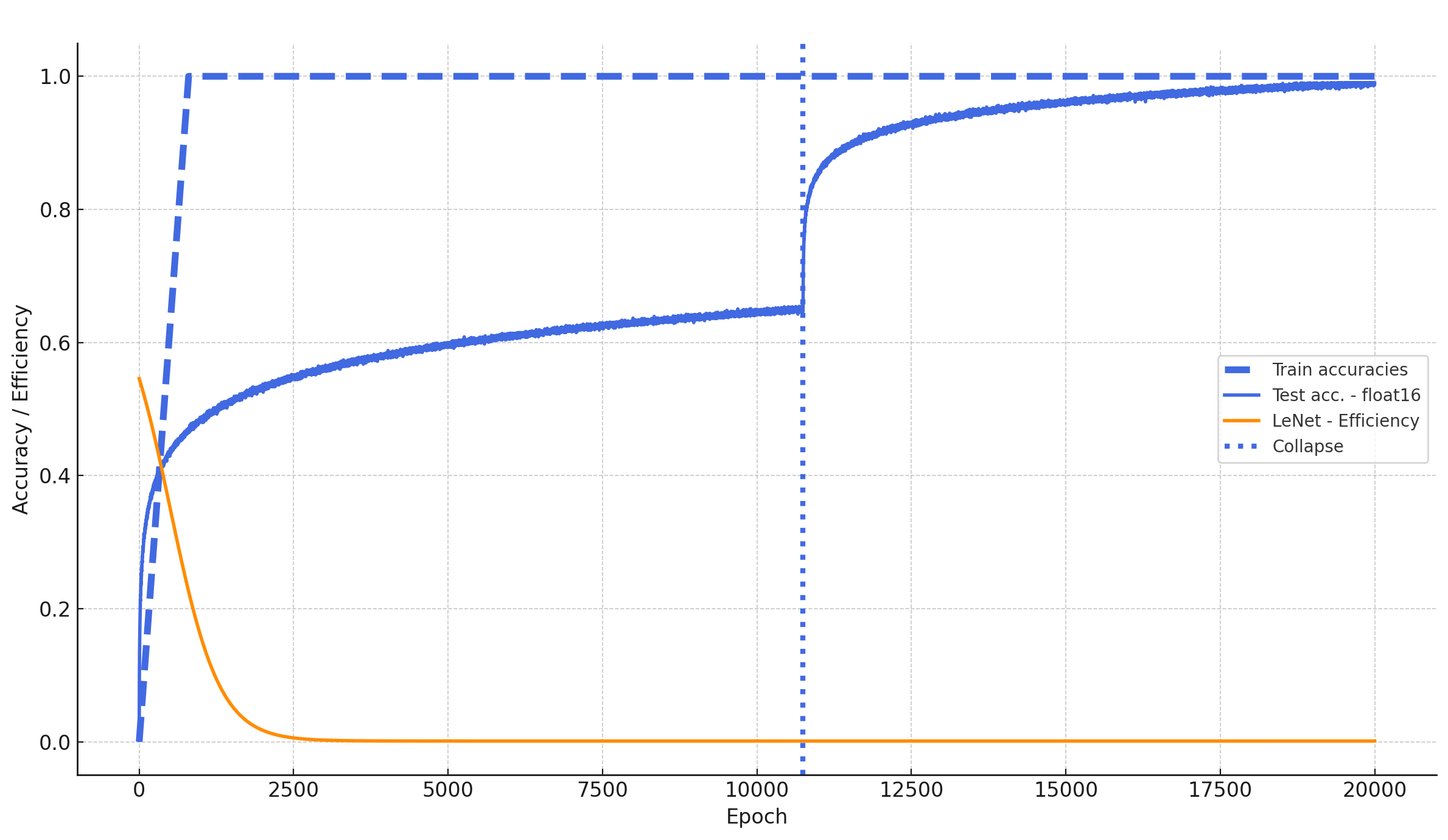}
        \caption{LeNet 16-bit}
        \label{fig:lenet_16}
    \end{subfigure}
    \hfill
    \begin{subfigure}[b]{0.32\textwidth}
        \includegraphics[width=\textwidth]{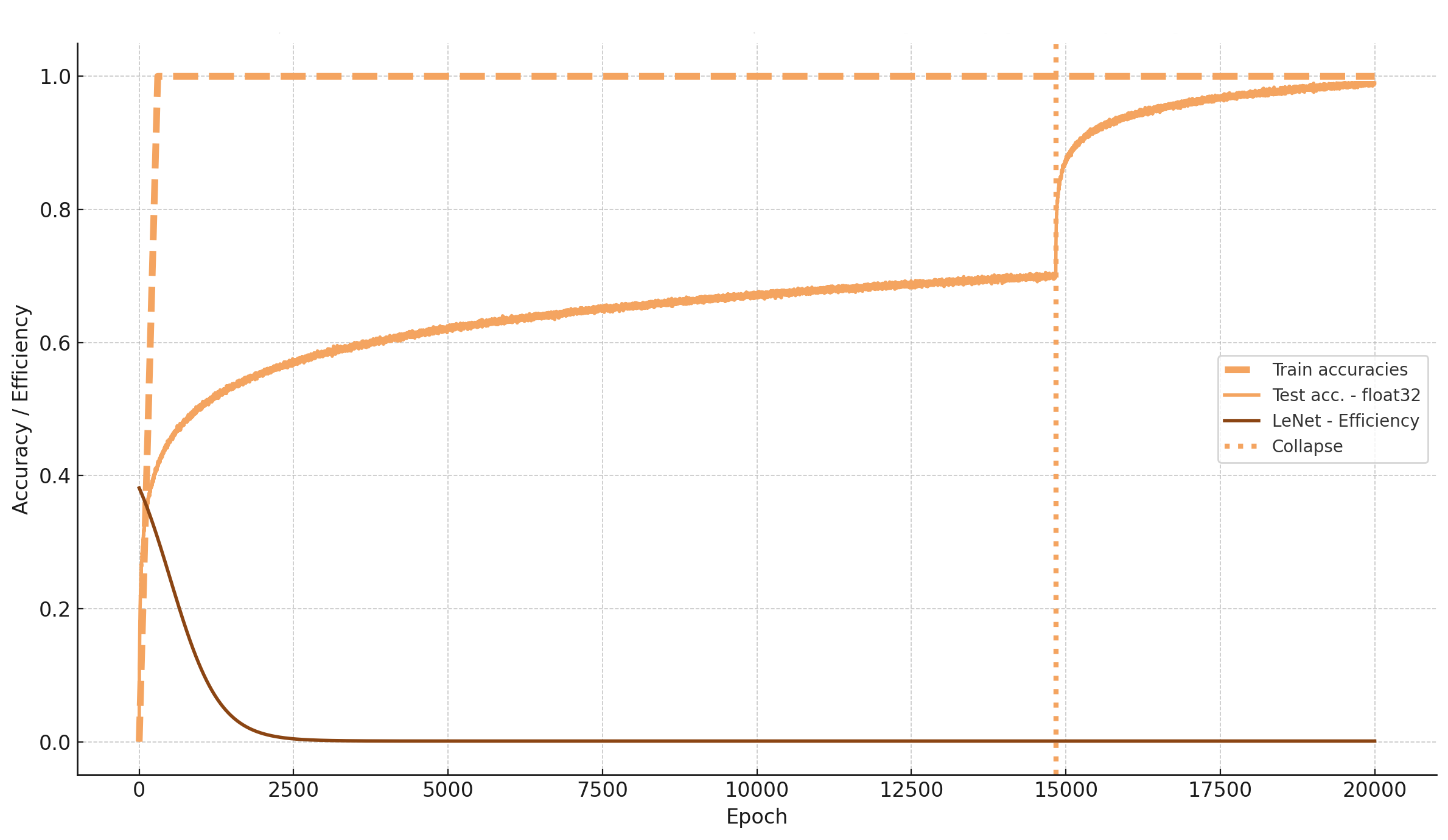}
        \caption{LeNet 32-bit}
        \label{fig:lenet_32}
    \end{subfigure}
    \hfill
    \begin{subfigure}[b]{0.32\textwidth}
        \includegraphics[width=\textwidth]{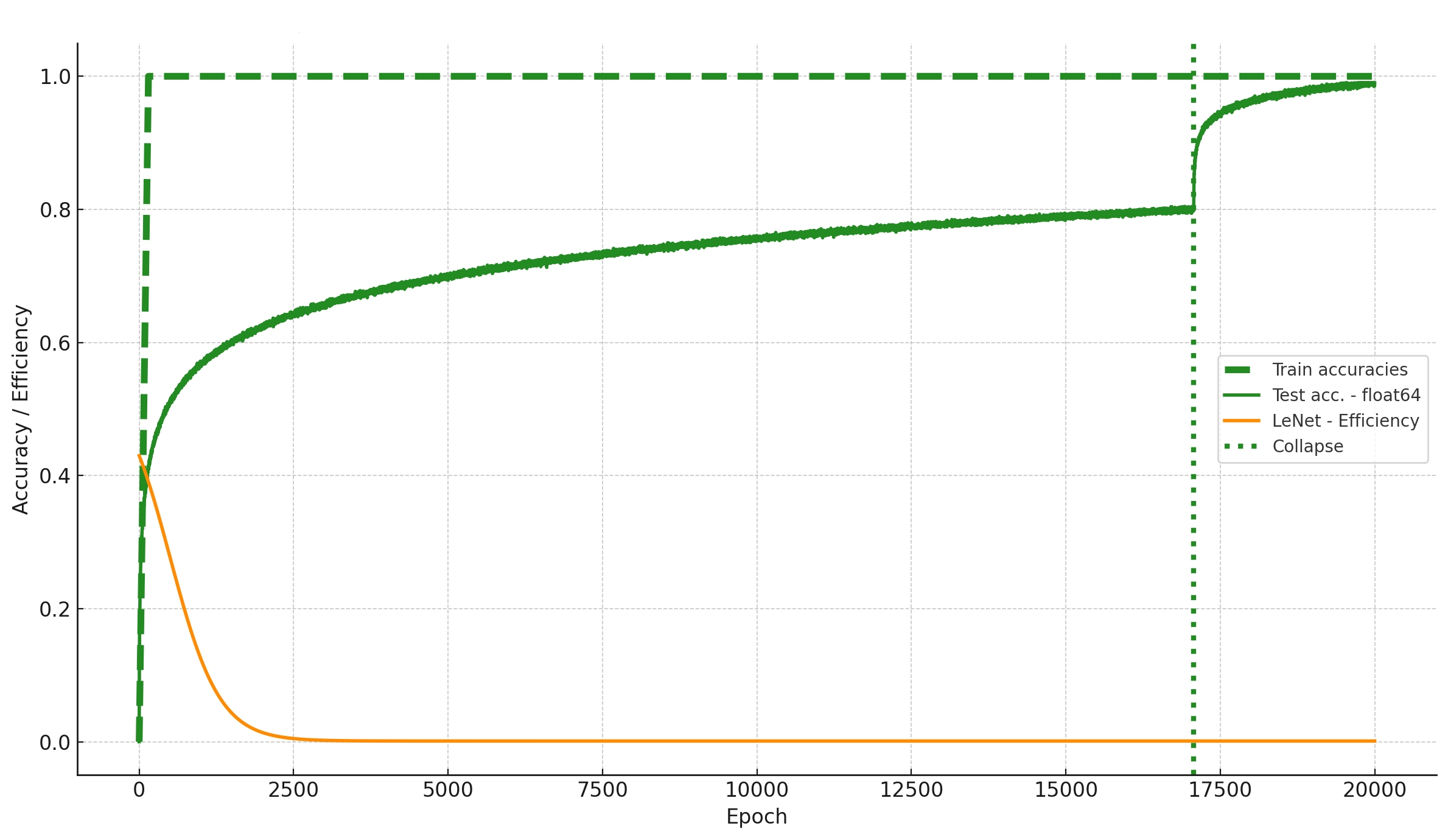}
        \caption{LeNet 64-bit}
        \label{fig:lenet_64}
    \end{subfigure}
    \caption{The performance and efficiency per epoch for a LeNet CNN, using the MNIST dataset, using a 16,32 and 64-bit variable precision.}
    \label{fig:lenet_long}
\end{figure*}

\begin{figure*}[t]
    \centering
    \begin{subfigure}[b]{0.32\textwidth}
        \includegraphics[width=\textwidth]{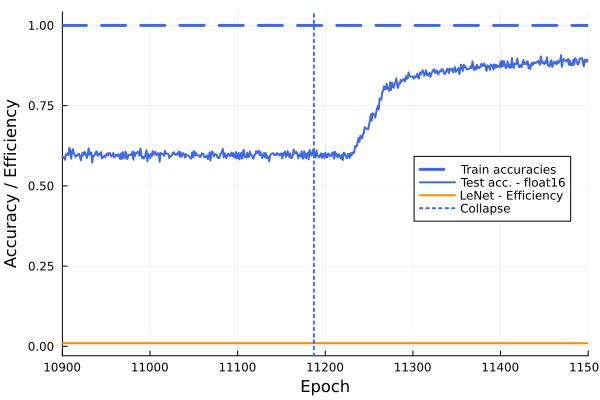}
        \caption{LeNet 16-bit zoomed}
        \label{fig:lenet_16_zoom}
    \end{subfigure}
    \hfill
    \begin{subfigure}[b]{0.32\textwidth}
        \includegraphics[width=\textwidth]{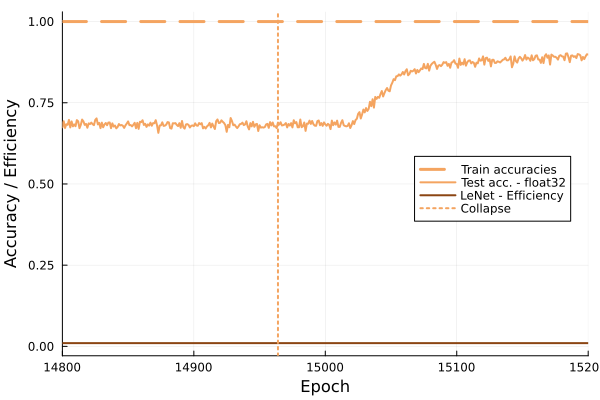}
        \caption{LeNet 32-bit zoomed}
        \label{fig:lenet_32_zoom}
    \end{subfigure}
    \hfill
    \begin{subfigure}[b]{0.32\textwidth}
        \includegraphics[width=\textwidth]{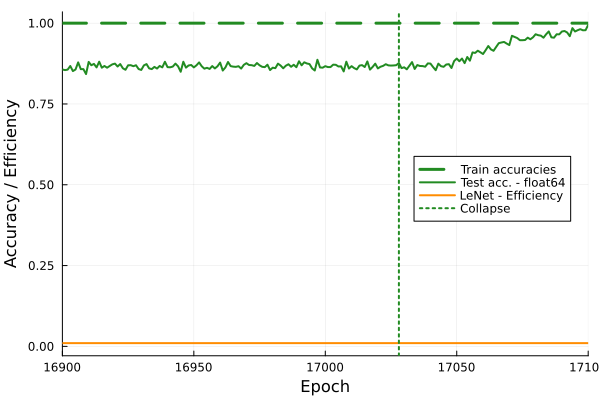}
        \caption{LeNet 64-bit zoomed}
        \label{fig:lenet_64_zoom}
    \end{subfigure}
    \caption{Zoom on the region where the orthograd algorithm is applied, for a LeNet CNN using the MNIST dataset, at 16, 32 and 64-bit variable precision.}
    \label{fig:lenet_long_zoom}
\end{figure*}

\section{Tabular Data for Section~\ref{sec:shorttrainingtime}}
For this section we provide a table (see Table~\ref{tab:grokking_numerical_results_mean_quick}) containing the accuracy values near each application of the Orthograd algorithm, i.e., two epochs before and after $50$ ($47$, $48$, $51$, $52$); similarly for $100$ and $150$. These values are reported for each of the experiments show in  Figures~\ref{fig:percept_16_quick},~\ref{fig:percept_32_quick},~\ref{fig:percept_64_quick},~\ref{fig:lenet_16_quick},~\ref{fig:lenet_32_quick}, and~\ref{fig:lenet_64_quick}.

\begin{table*}[t]
  \centering
  \caption{Numerical results for the early-time grokking experiments. The leftmost column shows the model type MLP that performed the modulo 113 task; LeNet that performed the MNIST task. The columns accuracy value -2 and -1 record the accuracy two and one epoch before applying the Orthograd algorithm, respectively. Similarly, accuracy value +1 and +2 records the accuracy one and two epochs after applying the Orthograd algorithm, respectively.}
  \label{tab:grokking_numerical_results_mean_quick}
  \footnotesize
  \setlength{\tabcolsep}{3pt}
  \begin{tabular}{lrrrrr}
    \toprule
    & \textbf{Epoch} & \textbf{Accuracy $-2$} & \textbf{Accuracy $-1$} & \textbf{Accuracy $+1$} & \textbf{Accuracy $+2$} \\
    \midrule
    MLP 16-bit & 50 & 0.3198 & 0.3208 & 0.3228 & 0.3235 \\
    MLP 16-bit & 100 & 0.3496 & 0.3516 & 0.3516 & 0.3517 \\
    MLP 16-bit & 150 & 0.3506 & 0.3503 & 0.3516 & 0.3520 \\
    MLP 32-bit & 50 & 0.31047714 & 0.31050083 & 0.31050990 & 0.3105144 \\
    MLP 32-bit & 100 & 0.36048922 & 0.36049768 & 0.36049932 & 0.36050236 \\
    MLP 32-bit & 150 & 0.40048927 & 0.40049290 & 0.40050557 & 0.40051800 \\
    MLP 64-bit & 50 & 0.31098664593165456 & 0.31099036016668750 & 0.31100269995742355 & 0.31101400602066680 \\
    MLP 64-bit & 100 & 0.34098711699073030 & 0.34099176252251040 & 0.34100344711960310 & 0.34101115044821220 \\
    MLP 64-bit & 150 & 0.34094213766241880 & 0.34094782313983835 & 0.34095876696574884 & 0.34096879802585295 \\
    \addlinespace
    LeNet 16-bit & 50 & 0.3005 & 0.3013 & 0.3013 & 0.3015 \\
    LeNet 16-bit & 100 & 0.3691 & 0.3699 & 0.3704 & 0.3718 \\
    LeNet 16-bit & 150 & 0.4097 & 0.4110 & 0.4111 & 0.4119 \\
    LeNet 32-bit & 50 & 0.36098915 & 0.36099428 & 0.36100575 & 0.36102572 \\
    LeNet 32-bit & 100 & 0.41099543 & 0.41099980 & 0.41100586 & 0.41101025 \\
    LeNet 32-bit & 150 & 0.42098250 & 0.42100827 & 0.42100880 & 0.42103056 \\
    LeNet 64-bit & 50 & 0.40099838981655235 & 0.40100167728465540 & 0.40100441104333820 & 0.41001773879209060 \\
    LeNet 64-bit & 100 & 0.47048626152014684 & 0.47049831799038690 & 0.47050874309248315 & 0.47351139685790770 \\
    LeNet 64-bit & 150 & 0.50149274260563670 & 0.50251099164355230 & 0.50151271764247890 & 0.50252116270772444 \\
    \bottomrule
  \end{tabular}
\end{table*}

\section{Tabular Data for Section~\ref{sec:longtrainingtime}}

In this section we provide a table (see Table~\ref{tab:grokking_numerical_results_mean}) containing the mean Orthograd-application epochs and the corresponding mean grokking epochs for each of the experiments shown in  Figures~\ref{fig:percept_16},~\ref{fig:percept_32},~\ref{fig:percept_64},~\ref{fig:lenet_16},~\ref{fig:lenet_32}, and~\ref{fig:lenet_64}.

\begin{table}[t]
  \centering
  \caption{Results for the late-time grokking experiments (averaged over 3 runs).}
  \label{tab:grokking_numerical_results_mean}
  \small
  \setlength{\tabcolsep}{4pt}
  \begin{tabular}{lrr}
    \toprule
    & \textbf{Orthograd epoch} & \textbf{Grokking epoch} \\
    \midrule
    MLP 16-bit & 12287 & 12329 \\
    MLP 32-bit & 9731 & 9775 \\
    MLP 64-bit & 7484 & 7523 \\
    \addlinespace
    LeNet 16-bit & 11193 & 11252 \\
    LeNet 32-bit & 14964 & 15041 \\
    LeNet 64-bit & 17046 & 17105 \\
    \bottomrule
  \end{tabular}
\end{table}

\section{Formal Proof} \label{sec:proof}

This section examines the conditions under which training efficiency is monotonically decreasing over training epochs, including steps at which accuracy increases abruptly. We consider the generalized efficiency ratio model
\begin{align}\label{eq:efficiency}
    E_t = \eff(\bar{x},y,i=t) &= f(x_t,y_t) \\
                              &= \frac{x_t}{y_t}, \nonumber \\
                              x_t\in[0,1],&\; y_t\in(0,\infty). \nonumber
\end{align}
Where $t\in\mathbb{N}$ indexes epochs, $x_t$ represents the accuracy quantity used in the numerator, and $y_t$ represents cumulative energy consumption. When energy is cumulative, we also impose $y_{t+1}\geq y_t$. Only epochs with positive cumulative energy are included in $E_t$.

The arguments below concern Equation~\eqref{eq:efficiency}. Applying them to an efficiency expression involving training accuracy, test accuracy, and other variables requires an explicit identification of its numerator and denominator with $x_t$ and $y_t$.

\begin{definition}[Monotonic decrease]
The sequence $(E_t)_{t\in\mathbb{N}}$ is non-increasing if $E_{t+1}\leq E_t \;\text{for every }t\in\mathbb{N}.$ Equivalently, $E_s\geq E_t$ whenever $s\leq t$. Throughout this section, monotonically decreasing means non-increasing; equality between successive values is permitted.
\end{definition}

\begin{proposition}[Monotonicity in the denominator]\label{prop:denominator}
For any fixed $x\in[0,1]$ and $0<y_1\leq y_2$, $f(x,y_2)\leq f(x,y_1)$.
\end{proposition}
\begin{proof}
Positivity gives $1/y_2\leq 1/y_1$. Multiplying by $x\geq0$ preserves the inequality and yields $x/y_2\leq x/y_1$.
\end{proof}

Proposition~\ref{prop:denominator} holds with the numerator fixed. It does not establish monotonic decrease over time when the numerator and denominator both change. A separate condition on their changes is required.

We propose a minimal increase in energy described by \[y_t,y_{t+1}>0,\qquad |y_t-y_{t+1}|\leq0.\] Since an absolute value is nonnegative, this condition implies $y_t=y_{t+1}$, rather than merely $y_t\leq y_{t+1}$.

\begin{theorem}[Exact condition for monotonic decrease]\label{thm:condition}
Let $(x_t)_{t\in\mathbb{N}}$ and $(y_t)_{t\in\mathbb{N}}$ satisfy $x_t\in[0,1]$ and $y_t>0$ for every $t$. Define $E_t=x_t/y_t$. Then $(E_t)$ is non-increasing if and only if
\begin{equation}\label{eq:cross-product}
    x_{t+1}y_t\leq x_ty_{t+1}
    \qquad\text{for every }t\in\mathbb{N}.
\end{equation}
\end{theorem}
\begin{proof}
Fix an arbitrary time $t$. Since $y_ty_{t+1}>0$, multiplication by this product preserves inequalities. Hence
\begin{align*}
    E_{t+1}\leq E_t
    &\iff \frac{x_{t+1}}{y_{t+1}}\leq\frac{x_t}{y_t} \\
    &\iff x_{t+1}y_t\leq x_ty_{t+1}.
\end{align*}
Applying this equivalence at every time proves both directions. Furthermore, if $s<t$, repeated application of the adjacent-step inequalities gives $E_t\leq E_{t-1}\leq\cdots\leq E_s$. For $s=t$ the required inequality is equality.
\end{proof}
Theorem~\ref{thm:condition} characterizes the additional condition needed. It does not establish that a training process satisfies that condition; such a conclusion requires separate assumptions or results about the accuracy and energy sequences.
\begin{corollary}[Comparison of changes]\label{cor:changes}
Define $\Delta x_t=x_{t+1}-x_t$ and $\Delta y_t=y_{t+1}-y_t$. Under the assumptions of Theorem~\ref{thm:condition},
\begin{equation}\label{eq:changes}
    E_{t+1}\leq E_t
    \iff y_t\Delta x_t\leq x_t\Delta y_t.
\end{equation}
If additionally $x_t>0$, this is equivalent to
\begin{equation}\label{eq:relative}
    \frac{\Delta x_t}{x_t}\leq\frac{\Delta y_t}{y_t}.
\end{equation}
\end{corollary}
\begin{proof}
Substitute $x_{t+1}=x_t+\Delta x_t$ and $y_{t+1}=y_t+\Delta y_t$ into Equation~\eqref{eq:cross-product}. Canceling $x_ty_t$ gives Equation~\eqref{eq:changes}. When $x_t>0$, division by $x_ty_t>0$ yields Equation~\eqref{eq:relative}.
\end{proof}
For positive accuracy, the proportional increase in cumulative energy must therefore be at least as large as the proportional increase in accuracy. The undivided condition~\eqref{eq:changes} also covers $x_t=0$. In that case, non-increasing efficiency requires $x_{t+1}=0$.
\begin{corollary}[A sufficient condition with non-increasing accuracy]
If $0\leq x_{t+1}\leq x_t$ and $0<y_t\leq y_{t+1}$ at every step, then efficiency is non-increasing.
\end{corollary}
\begin{proof}
For every $t$,
\[
    \frac{x_{t+1}}{y_{t+1}}
    \leq\frac{x_t}{y_{t+1}}
    \leq\frac{x_t}{y_t}.
\]
The first inequality uses the ordering of the numerators and positivity of $y_{t+1}$; the second uses Proposition~\ref{prop:denominator}.
\end{proof}
If grokking produces an abrupt increase in the accuracy quantity $x_t$ used by this model, efficiency can either increase or decrease. For example, a change from $x_t=0.2$ to $x_{t+1}=0.8$ multiplies accuracy by four. Efficiency is non-increasing at this step exactly when $y_{t+1}\geq4y_t$. Equality leaves efficiency unchanged; a larger energy ratio makes it decrease strictly. Accordingly, a justified statement for our efficiency model is:
\begin{quote}
For positive cumulative energy, training efficiency is non-increasing provided that $x_{t+1}y_t\leq x_ty_{t+1}$ at every training step, including steps at which grokking occurs.
\end{quote}
To claim that efficiency always decreases regardless of grokking, this condition must be derived from several training process measurements. Verifying it on measured trajectories establishes the property for those trajectories, but does not prove it for all possible training runs. A computational proof our statements using LEAN theorem prover is provided at \url{-}.

\section{Analysis and Discussion}

The research question of this paper was whether the monotonic decrease property of the training efficiency function holds at any point during training in the presence of grokking. Our experimental results indicate that yes training efficiency does hold the monotonic property even if grokking occurs, and that this holds across multiple model architectures (LeNet CNNs and MLP), tasks and through different variable precisions. A potential contributing factor to this is that grokking occurs so late in training that the cumulative sum of efficiency terms is so large in comparison to the accuracy gains that these gains are insignificant in the overall calculations of efficiency. In this context our findings from the experiments in Section \ref{sec:shorttrainingtime} are important because they provide empirical evidence that indicate that grokking cannot occur early in a training process.

Additionally, the comparative analysis between different architectures such as the Multi-Layer Perceptron (MLP) and LeNet Convolutional Neural Network (CNN) highlights intriguing architectural dependencies in grokking behavior. CNN architectures tend to start growing closer to the collapse point\footnote{Figures~\ref{fig:lenet_16_zoom},~\ref{fig:lenet_32_zoom} and~\ref{fig:lenet_64_zoom}.}, whereas MLPs display variability in the grokking start and intensity. In other words, CCN models appear more sensitive to the orthograd model then MLPs, and this is a potential basis for future research on grokking. 

\paragraph{Limitations.} Our efficiency proxy depends on hardware and implementation details; thus further experiments with different learning tasks and across computing systems are recommended \citep{pomarico_grokking_2025,prakash_grokking_2025,prieto_grokking_2025,demoss_complexity_2024}.

\section{Conclusion}

In conclusion, this paper has provided empirical evidence that shows that the grokking phenomena does not affect the general dynamics of efficiency in neural network training processes. Furthermore, our results show that early perturbations (collapse) of the loss function does not induce grokking nor modify the dynamics of efficiency. Consequently, this paper has provided evidence that overfitting is a requirement for grokking to appear on a training regime, as no grokking happens at early stages of training the behavior of efficiency is not modified. To further reinforce the fact that the $\eff(\cdot)$ will always be stable under grokking a mathematical proof of the statement is included in Section~\ref{sec:proof}.

By systematically identifying and analyzing grokking events across different architectures and numerical precisions, the thesis demonstrates that efficiency is stable through grokking. More specifically, Equation~\ref{eq:efficiencygrokking} is stable, through sudden increases in testing accuracy, which happen when a Neural Network groks. We hypothesize that the reason why grokking does not cause an increase in efficiency at later stages in training is that grokking typically occurs so late on in the training process that the accumulated energy cost dominates the efficiency calculation and even large increases in accuracy cannot overcome the accumulated cost.

Similarly, our experiments also provide evidence showing that grokking is not present at early stages of training (up 500 epochs) even when applying the state-of-the-art technique of Softmax Collapse, and this further support the argument that training efficiency is a monotonically decreasing function through all the training of a DNN, not only through early stages of training, but also at later stages of training (up to 20,000 epochs).

%This should be the final version of this paragraph but it was commented
% so we remain anonymous (autors -> relevant works).
%The experimental challenges encountered in terms of numerical instability across long training runs foregrounded the need to adjust and calibrate the energy sampling rate during extremely long training runs in order to avoid underflow and overflow errors when using the efficiency measurement methodology proposed by~\citeauthor{cueto-mendoza_framework_2024}. These challenges point to the need for the development of architectural and numerical strategies that enhance the robustness of the efficiency framework against numerical instability in future research. However, the next chapter of this thesis develops the efficiency methodology in a different direction.

The experimental challenges encountered in terms of numerical instability across long training runs foregrounded the need to adjust and calibrate the energy sampling rate during extremely long training runs in order to avoid underflow and overflow errors when using the efficiency measurement methodology proposed by relevant works. These challenges point to the need for the development of architectural and numerical strategies that enhance the robustness of the efficiency framework against numerical instability in future research. However, the next chapter of this thesis develops the efficiency methodology in a different direction.

\subsection*{AI use statement}

In this work, we used generative AI tools for the preparation of the manuscript, specifically for assistance with \LaTeX{} formatting of the figure environments in Section~\ref{ssec:results}. We have not used generative AI tools for research ideation, experimental design, the implementation or execution of the experiments, the analysis of the resulting data, or the derivation of the proof in Section~\ref{sec:proof} with the exception of the provided LEAN4 proof that was assisted by AI; the substantive text of this work was written by the authors. Retrieval, translation and data-collection uses of generative AI are not applicable to this work. We have reviewed all AI-assisted work. Concretely, the AI-assisted changes were limited to the layout of existing figures, namely the choice of float environment, panel widths and caption placement; the figures themselves, their underlying data and their captions' content were produced by the authors, and every cross-reference and the compiled output were checked by the authors against the source data. We take responsibility for the final content of this work, including text, claims, or artifacts produced with the aid of generative AI.

\subsubsection*{Acknowledgments}
All acknowledgments go at the end of the paper, including thanks to reviewers who gave useful comments, to colleagues who contributed to the ideas, and to funding agencies and corporate sponsors that provided financial support.
To preserve the anonymity, please include acknowledgments \emph{only} in the camera-ready papers. The acknowledgments do not count against the 9-page page limit in the camera-ready.
%This work was conducted with the financial support of the Research Ireland Centre for Research Training in Digitally-Enhanced Reality (d-real) under Grant No. 18/CRT/6224. For the purpose of Open Access, the author has applied a CC BY public copyright licence to any Author Accepted Manuscript version arising from this submission.

\subsubsection*{References}

\bibliographystyle{unsrtnat}
\bibliography{references}

%\clearpage
%\appendix
%\thispagestyle{empty}

\end{document}